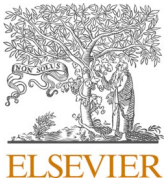
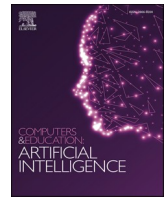

# LLaVA-docent: Instruction tuning with multimodal large language model to support art appreciation education

Unggi Lee [a,*], Minji Jeon [b], Yunseo Lee [c,1], Gyuri Byun [d,1], Yoorim Son [e,1], Jaeyoon Shin [f,1], Hongkyu Ko [f,**], Hyeoncheol Kim [a,***]

[a] *Department of Computer Science and Engineering, Korea University, South Korea*
[b] *Teaching, Learning and Teacher Education, University of Nebraska-Lincoln, United States*
[c] *Poongnap Elementary School, Seoul Metropolitan Office of Education, South Korea*
[d] *Department of Education, Seoul National University, South Korea*
[e] *Interdisciplinary Program in Art Education (Art Education Major), Seoul National University, South Korea*
[f] *Department of Elementary Art Education, Seoul National University of Education, South Korea*

## ARTICLE INFO

## ABSTRACT

Despite the development of various AI systems to support learning in various domains, AI assistance for art appreciation education has not been extensively explored. Art appreciation, often perceived as an unfamiliar and challenging endeavor for most students, can be more accessible with a generative AI enabled conversation partner that provides tailored questions and encourages the audience to deeply appreciate artwork. This study explores the application of multimodal large language models (MLLMs) in art appreciation education, with a focus on developing LLaVA-Docent, a model designed to serve as a personal tutor for art appreciation. Our approach involved design and development research, focusing on iterative enhancement to design and develop the application to produce a functional MLLM-enabled chatbot along with a data design framework for art appreciation education. To that end, we established a virtual dialogue dataset that was generated by GPT-4, which was instrumental in training our MLLM, LLaVA-Docent. The performance of LLaVA-Docent was evaluated by benchmarking it against alternative settings and revealed its distinct strengths and weaknesses. Our findings highlight the efficacy of the MMLM-based personalized art appreciation chatbot and demonstrate its applicability for a novel approach in which art appreciation is taught and experienced.

## 1. Introduction

Appreciating works of art is often perceived as an elite activity reserved for a select group of interested individuals (DiMaggio & Useem, 1978; Ostrower, 1998), causing many people to feel intimidated or lack confidence when encountering art in daily lives. The absence of effective guidance and education for art appreciation in art classes and museums hampers deeper engagement with art (Duh et al., 2012). Many viewers, though enthusiastic about interpreting artworks, often pass by them with only a cursory glance due to insufficient appreciation guidance and training (Deeth, 2015). Novice or inexperienced art viewers, whether in a museum or classroom, often depend on information provided by curators and teachers, but this information rarely encourages personal engagement with the art. It has been argued that museums and educational institutions should adopt constructivist theory to help viewers connect artworks to their personal lives (Hein, 1999). In response, many museums are reforming their exhibition styles and offering more verbal and practical activities for the audience (Deeth, 2015). Nevertheless, these solutions remain challenging in K-12 settings, where engaging and interactive discussion is limited due to practical reasons, for example, outdated teaching methods, high student-to-teacher ratio, and diminished interest in learners who seldom receive positive feedback (Li, 2020). Providing immediate and personalized feedback to each student, who may have diverse learning capabilities and paces (Hayadi et al., 2018; Moubayed et al., 2020), considerably increases educator's workloads, resulting in fewer opportunities for deeper engagement with art. Despite these challenges, close art appreciation education during the K-12 years is important, as it helps cultivate the attitudes and skills necessary to enjoy and confidently appreciate art later in life (Sickler-Voigt, 2019).

*1.1. Landscape of technology-enabled art education and potential of AI application*

Digital art museums, repositories, and websites, such as Google Arts & Culture, may be considered as a way to address the lack of engagement with art and interaction as they have reduced geographical, economic, and cultural barriers and allowed to access a vast array of artworks online without time constraints (Verde & Valero, 2021; Wahyuningtyas, 2017). However, increased accessibility alone has not been sufficient to help novices engage deeply with art, as many still lack the skills to use these tools effectively. The ability to view works of art in detail using online platforms can be overwhelming for learners on their own (Lindong & Lim, 2025) due to cognitive overload from navigating the tools and not knowing where to start among the numerous available art pieces (Zhang, 2020). To foster art appreciation, K-12 environments need not only readily available archives or repositories but also pedagogy-based tools to effectively scaffold learners to analyze artworks, and develop their interpretations on their own. Given this, AI as an interactive learning assistant for art discussion can be beneficial as it can contribute to engage learners in reflective thinking, focused attention, and real-time interaction followed by immediate feedback which fosters more in-depth exploration and persists learning (Chen et al., 2020).

In light of the affordances of AI in various domains, many subject areas have made numerous attempts to overcome the lack of real-time scaffolds and interactions with more knowledgeable others through AI (Chang et al., 2021; Koć-Januchta et al., 2020; Nazari et al., 2021). However, AI's potential remains comparatively underutilized in art education (Chiu et al., 2022). In step with the image creation technology trends, AI tools for art expression education have evolved, primarily focusing on enhancing painting and creative skills by providing personalized feedback (e.g., DL-ALS; AI-CTSAM; Chiu et al., 2022; Fan & Zhong, 2022). For example, Midjourney, DALL-E, and Stable Diffusion are widely used across proficiency levels and age groups, in generating diverse output when human input, even small cues, is given (Derevyanko & Zalevska, 2023). On the other hand, art appreciation has fewer available AI tools than artistic expression. Art appreciation, with its inherently distinctive processes than expression, has unique needs and implications for developing an AI application. For example, while divergent thinking and idea generation could be key elements to be aimed for in AI for expression, AI tools for appreciation should aim to facilitate logical thinking and detailed observation. Therefore, it is crucial to develop appreciation-specific frameworks, enabling students to examine the elements and principles of design, providing historical context, and engaging them in discussion, thereby promoting critical and logical thinking. These art appreciation specific data design frameworks can be used to analyze nuanced conversations within AI tools and establish curated datasets for training AI models.

*1.2. AI-enabled assistant toward constructivist art pedagogy*

As demonstrated above, an AI tool for art appreciation education requires a carefully designed framework that guides the creation of datasets for training and analyzing human interactions with the AI assistant. With this, constructivist art pedagogy further enhances this AI development by emphasizing the roles of interpreters as well as the importance of meaning-making process in art appreciation. Together, constructivist pedagogy and AI affordances can be good companions and go hand in hand. First, a central tenet of constructivist approaches is scaffolding that provides learners with tailored support (Roehler & Cantlon, 1997). An interactive AI assistant can provide scaffolding by fostering personalized learning experiences. Constructivist approaches in art appreciation shifts the focus from seeking expert-recognized answers to encouraging personal interpretations, thus creating a more individualized and meaningful engagement with art. Pedagogically sound technologies, as we intended to develop, can realize this paradigm shift, leading to more principled art appreciation education.

On the other hand, developing such an interactive AI assistant requires incorporating coherent and effective design principles that inform the design of target functionalities to foster active interaction and deeper engagement with art. Previous research on art appreciation education offers insights for developing these capabilities. Key educational theories include Anderson's critical stages (1993), visual thinking strategies by Yenawine (2013), Arenas's conversation-oriented approach (Yoshida, 2009), and artful thinking (Tishman & Palmer, 2006). These pedagogical approaches emphasize learners' interpretation and interactive learning process, fostering a deeper understanding (Vygotsky, 1978). For example, *Anderson's stages* (1993) emphasize personal experience and reaction to art, involving stages such as *reaction, perceptual analysis, personal interpretation, contextual examination,* and *synthesis* (as illustrated in Table 5 and Appendix 3). *Visual thinking strategies*, developed by Yenawine (2013), utilizes inquiry and paraphrasing to activate students' observational skills. Arenas advocated for *dialogic interaction* in art appreciation, prioritizing personal observation over expert explanation (Kinoshita, 2001; Yoshida, 2009). *Artful thinking* employs targeted questioning to deepen students' cognitive and emotional engagement with art (Tishman & Palmer, 2006). As research indicates that individuals engage in more extensive and prolonged conversations with chatbots compared to human-human interaction (Hill & Farreras, 2015), learners may discuss artwork more freely and concretely through one-on-one interactions with AI and develop their knowledge and skill for appreciating art. Personalized, tailored contents and scaffolding strategies can foster independent, reflective appreciators who connect their everyday life to art. By integrating these constructivist principles, an AI-enabled assistant can enhance art appreciation experience, making it more relevant, engaging, dynamic, and hence effective.

*1.3. Open-source MLLM as a breakthrough for constructivist art appreciation education*

Recent technological advancements have attempted to incorporate scaffolding in their solutions for learning through Large Language Models (LLMs) like GPT-4 and Claude-3 Opus. With their generative capability from web-scale data training, LLMs can perform cognitive tasks rivaling human ability, which can devise and implement appropriate scaffolding if model prompts are provided. However, conventional LLMs lack visual modality, crucial in art education. Multimodal Large Language Models (MLLMs) like GPT-4o and Gemini-pro have emerged to address this limitation, with their visual processing capabilities alongside text and enabled a more comprehensive approach to art appreciation. This aligns with the constructivist emphasis on interpretation and meaning-making in visual contexts.

Another drawback of conventional LLMs is that they were developed by major tech companies and have limited scope and scalability. Therefore, when applied to specific educational contexts such as art appreciation, these closed models are difficult to optimize beyond basic prompt engineering, potentially resulting in conversations that fail to align with or adequately support intended educational objectives. Further, using these models necessarily requires sharing learner data with big tech companies, which might raise privacy concerns. Lastly, these models typically require stable internet access, limiting their use in environments with limited or no connectivity.

To leverage the strengths of LLMs while mitigating these concerns, open-source models present a promising alternative. These models can maintain robust performance while allowing for customization and on-device deployment. With these open-source models, AI tools can be developed, tailored specifically for art appreciation education, which ensures more task-oriented, focused interactions with enhanced privacy. Open-source models also improve accessibility in various environments, potentially overcoming the connectivity limitations inherent in closed models. In light of these considerations, this study explored the potential of open-source MLLMs for art appreciation education.

Specifically, we intended to develop LLaVA-Docent, a model based on the LLaVA [Large Language and Vision Assistant] framework (Liu, Li, Li, & Lee, 2023). LLaVA-Docent was designed to enhance interactive and personalized learning experiences in art appreciation overcoming the limitations of closed, proprietary models. The development of LLaVA-Docent involved a comprehensive data design framework that incorporates various attributes of exemplary artworks, pedagogical principles for art appreciation, and natural interaction. Therefore, our key research questions center on developing LLaVA-Docent and evaluating its potential to enhance art appreciation education. This involves considerations of the technological capabilities of these models as well as their potential practical application in diverse educational settings. Additionally, the study aims to propose methodologies for testing and evaluating the impact of LLaVA-Docent in real-world educational environments. Therefore, this study intends to investigate the following research questions: (a) What are the optimal characteristics of the data design framework, dataset, and model architecture for LLaVA-Docent in the context of art appreciation education, particularly considering distinct phases of art appreciation? (b) How does LLaVA-Docent demonstrate conversation quality for art appreciation education, as evaluated through both qualitative and quantitative measures? (c) What are the broader implications and potential applications of using open-source MLLMs like LLaVA-Docent in art appreciation education and similar educational contexts?

## 2. Method

### 2.1. Research design & procedure

This study concentrated on the iterative enhancement of the LLaVA model for art education, according to the design and development research (DDR) methodology Type 1 (Richey & Klein, 2014), which focuses on the systematic study and improvement of specific educational products or tools. As DDR Type 1 is essentialized with prototyping of a tool and the following validation, usability, field test, this study is involved with six phases (see Fig. 1). The process began with the development of the initial LLaVA-Docent Version 1 prototype in Phase 1, corresponding to the planning stage of DDR. Phase 2 encompassed a comprehensive literature review on the domain, art appreciation education and the validation process of the first prototype through initial interviews with subject matter experts (SMEs) (Table 1). This led to the development of *Data Design Framework Version 1*, reflecting the DDR stages of gathering information and developing a tentative product design. Specifically, for the validation of the tool, experts were chosen for their knowledge and experience in the domain area, art appreciation and art education, as well as their interest in AI applications (for the SME profiles, see Table 1). Semi-structured interviews were conducted, followed by thematic analysis (Braun & Clarke, 2006), which was used to develop and refine the *Data Design Framework Version 2* in the next phases. In Phase 3, the *Data Design Framework Version 1* was validated through additional SME interviews and refined as *Data Design Framework Version 2*, which correspond to the DDR steps of product revision. Phase 4 focused on data generation based on the *Data Design Framework Version 2*, corresponding to the DDR stage of developing the research product. In Phase 5, the training of LLaVA-Docent Version2 using the generated dataset, corresponding to the main product revision in DDR. Finally, Phase 6 incorporated quantitative and qualitative evaluations of LLaVA-Docent Version2, aligning with the DDR stages of main field testing and operational product revision.

### 2.2. Technological research development for LLaVA

#### 2.2.1. Architecture of LLaVA-Docent

In this study, LLaVA-Docentused LLaVA version 1.0. (Liu et al., 2023a, 2023b), consisting of three parts: vision encoder $g(\cdot)$, LLM $f_\phi(\cdot)$, and projection layer $W$. First, the vision encoder $g(\cdot)$ is specialized to capture the visual modality. The algorithm accepts image data $X_v$ in the vision encoder as input and returns $Z_v$ as output.

Second, a single trainable projection layer $W$ is used to link the different modalities between image and language. Projection layer $W$ projects visual feature $Z_v$ into visual embedding tokens $H_v$ with the same dimensionality of language embedding tokens $H_q$. The equation is $H_v = W \cdot Z_v$. Third, we choose Vicuna (Chiang et al., 2023) as LLM $f_\phi(\cdot)$, which is parameterized by $\phi$. Input data of LLM is concatenated $H_v$ and $H_q$. The final equation is $X_a = f_\phi(H_v \oplus H_q)$, which $\oplus$ is notation of concatenation. Fig. 2 is the architecture of LLaVA-Docent.

**Table 1**
Profiles of the SMEs.

| Phase | Expert | Occupation | Years of Experience |
|---|---|---|---|
| Phase 2 - Build Data Design Framework | Expert 1 | Elementary School Teacher | 5 years |
| | Expert 2 | Art Education Professor | 15 years |
| | Expert 3 | Media Artist | 11 years |
| | Expert 4 | Curator | 4 years |
| Phase 3 - Validate and refine Data Design Framework | Expert 5 | Art Education Professor | 5 years |
| | Expert 6 | High School Teacher | 11 years |

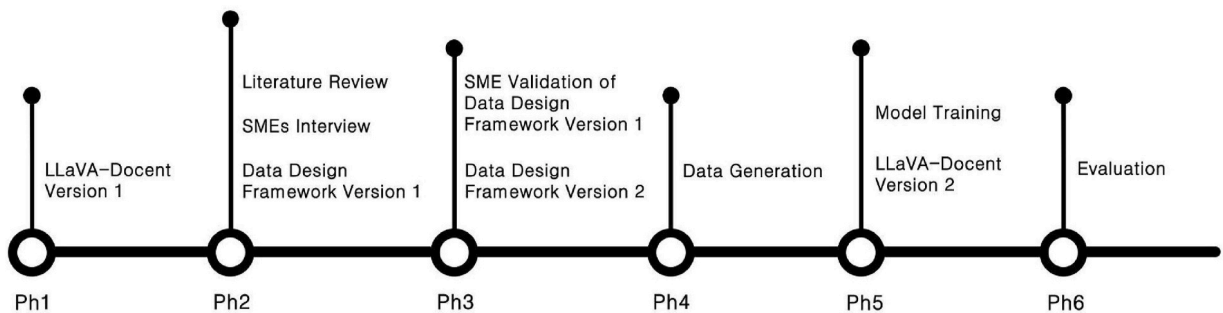

**Fig. 1.** Timeline of the research process.

### 2.2.2. Training strategies of LLaVA

Liu et al. (2023a, 2023b) suggested two-stage training strategies, pre-training and fine-tuning, to train LLaVA. In the pre-training stage, pairs of image and language data which explain the whole image and objects in the image, are used to train projection layer *W*. The projection layer *W* is trainable, while vision encoder $g(\cdot)$ and language model $f_\phi(\cdot)$ are not trainable. After the pre-training stage, projection layer *W* can link the visual and textual modality.

In the fine-tuning stage, pairs of image and dialogue data, which were generated from GPT-4, were used to fine-tune the pre-trained model. To generate dialogue data, we used a data design framework to make prompts for GPT-4 (section 4.4). In accordance with LLaVA's training strategy (Liu et al., 2023a), we focused on fine-tuning the model using only the dialogue parts of the data, excluding the prefix components. This approach enhances the model's ability to generate appropriate responses to novel dialogues in the target domain, improving its generalization capabilities.

### 2.2.3. Experiment setting for model training

In the model setting, we used vicuna-13b-v1.5 (Chiang et al., 2023) as LLM, clip-vit-large-patch14 (Radford et al., 2021) as image encoder and linear layer for projection layer, for both LLaVA-Docent Version1 and Version2. The model setting and hyper-parameter setting for training in this research are summarized in Table 2.

### 2.2.4. Model evaluation

Evaluating LLaVA-Docent entailed both quantitative and qualitative methods. LLaVA-Docent's performance was compared to GPT-4's zero-shot and few-shot capabilities, using modified prompts from Appendix 2 for the few-shot. Traditional metrics like BLEU and perplexity were set aside due to their limitations in capturing the nuanced dynamics of generative tasks (Bommasani et al., 2023; Celikyilmaz et al., 2020; Pang et al., 2020). Our evaluation process involved six researchers leading dialogues with both LLaVA-Docent and GPT-4 models, across three trials, each with 20 dialogue turns, totaling 360 turns of dialogue for analysis. Outputs were quantitatively assessed using a rubric aligned with distinct phases of art appreciation, known as Anderson's critical stages (Anderson, 1993). Qualitatively, dialogues were analyzed using a coding scheme by two pairs of researchers (see Appendix 5). This dual approach aimed to comprehensively capture both the model's technical accuracy and the contextual relevance in art appreciation conversations.

**Table 2**
Hyper-parameter setting for training LLaVA-Docent.

|  | Pre-training | Fine-tuning |
| --- | --- | --- |
| LLM | vicuna-13b-v1.5 | |
| Vision Encoder | clip-vit-large-patch14 | |
| Projection Layer | Linear layer | |
| Deepspeed | Zero-3 Offload | |
| Epoch | 1 | |
| Device per batch size (train/valid) | 128/128 | 32/32 |
| Weight decay | 0 | |
| Warmup ratio | 0.3 | |
| Learning rate | 2e-5 | |
| Max length | 2048 | |

## 3. Result

### 3.1. Phase 1: first prototype of LLaVA-Docent

We developed a prototype through pre-training and fine-tuning. In the pre-training stage, we used vicuna-13b-v1.5 (Chiang et al., 2023) as LLM, clip-vit-large-patch14 (Radford et al., 2021) as image encoder and linear layer for the projection layer. The image-text dataset for pre-training was cc3m_595k_images (Liu, 2023a), which was used to train the original LLaVA (Liu et al., 2023b). In the fine-tuning stage, the model setting was the same as the pre-training stage. The dataset for fine-tuning was LLaVA-Instruct-150K (Liu, 2023b), which included a virtual dialogue dataset generated from GPT-4.

### 3.2. Phase 2: Data Design Framework Version 1

Based on the prototype of LLaVA-Docent, we reviewed relevant literature on art criticism and art appreciation theories to establish a theoretical framework, and interviewed SMEs, the professionals in art appreciation education, which was led to develop a *Data Design Framework Version 1*. We found the theoretical background primarily on art criticism or art appreciation theories in instruction such as Anderson's critical stages (1993), visual thinking strategies by Yenawine (2013), Arenas's conversation-oriented approach (Yoshida, 2009), artful thinking (Tishman & Palmer, 2006). Anderson's (1993) theory states that the tool should address intrinsic and extrinsic information for appreciating artworks by following distinct phases of art appreciation, each of which is characterized with qualitatively different cognitive processes. Visual thinking strategies by Yenawine (2013) are based on constructivism, which demonstrated the tool should facilitate learning dialogues and interaction. Arenas provided concrete examples of dialogues of teachers and students in art appreciation classes (Yoshida, 2009), while artful thinking provides questioning skills to deepen students' engagements with art. These theories formed the basis of practical guidelines of the tool, which were embodied through the theoretical background of interactive learning assistant, LLaVA-Docent.

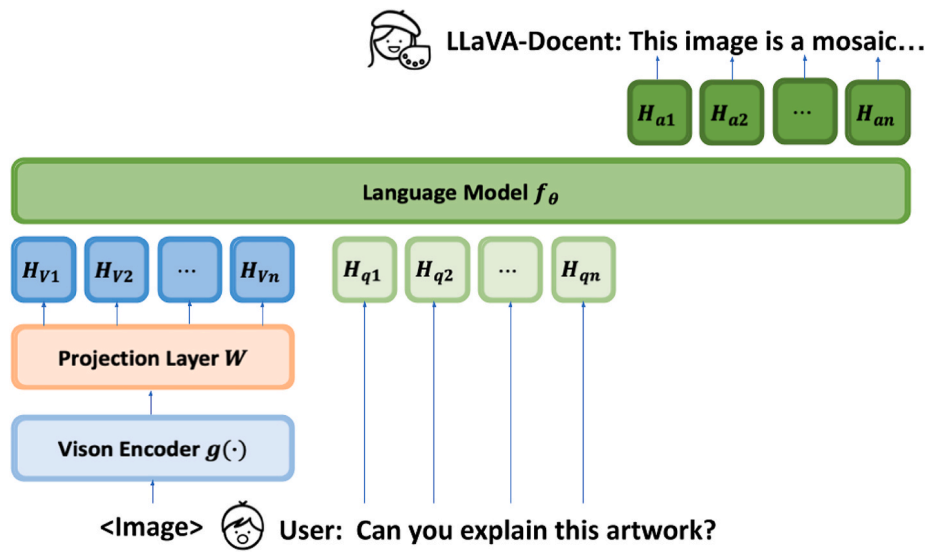

**Fig. 2.** Architecture of LLaVA-Docent, inspired by LLaVA (Liu et al., 2023b).

SME interviews offered practical ideas that suggested the types of information needed to interpret and appreciate artworks (data contents) and how such information would be represented and conveyed in dialogues (data forms), which resulted in becoming the cornerstones for the data design framework of the model (see Table 3). For example, intrinsic and extrinsic information of the artworks are included as the contents of the datasets, and multi-turn dialogues with open-ended questions are considered as the right form of the data. First, the data contents should be considered in the process of designing the dataset, with regards to intrinsic and extrinsic information of the artworks. Intrinsic information is what is inherent in the artwork itself, such as themes, figures, artistic style, colors, and positions, while extrinsic information refers to contextual cues, such as the artist's personal narratives, perspectives or styles, which gains more importance recently in interpreting artworks recently.

Second, prompts should be adjusted to be developmentally appropriate. Some SMEs are concerned that ordinary LLMs give excessively complicated and advanced contents about artwork, which needs to be more age appropriate for the youth. They claimed that the messages of this interactive app should be adjusted to the students' cognitive and affective levels, including plain language and encouraging prompts that help sustain their engagement. Also, they argued that LLaVA-Docent should select appropriate artworks, avoiding provocative or melancholic ones, which can be guidelines to make LLaVA-docent more effective in personalized art appreciation.

Lastly, data forms also should be designed to provide scaffolds based on the constructivist pedagogies. As all the SMEs provided suggestions on how the aforementioned content will be formulated and delivered to students, data forms were recommended to be designed to pose open-ended questions that could promote divergent thinking. Also, since this open-ended nature of the question may overwhelm students, this could be broken down into several prompts with multiple turn-taking. During these one-on-one interactions, students will be offered gradual scaffolds and tailored feedback, which elicit their personal engagement with the artwork.

### 3.3. Phase 3: validating the Data Design Framework Version1

To validate the initial iteration developing the distinct phases of art appreciation theoretical framework embedded in the data design framework, another SME interview was carried out. The experts in the SME interview comprised one professor specializing in art education and one high school art instructor. The SMEs evaluated the validity of *Data Design Framework Version 1*, including the theoretical framework, and offered further suggestions for improvement which is summarized as follows. Through this process, there were three confirmations and changes made and reflected in the *Data Design Framework Version 2*.

First, confirmation on the contents of the theoretical framework, including distinct phases for art appreciation, has been made. The first version of the theoretical framework, constructed through Phase 2, included the aligning perspectives of various scholars about art appreciation education. Based on this, the phases for art appreciation also involve integrated similar ideas and examples of art appreciation conversation that can be utilized and referred for educating individuals about art appreciation. In regards to this integrated approach, the SMEs expressed consensus on involving various conversation examples from various art educational theories while employing Anderson's art appreciation theory as the structure of the phases for art appreciation.

Second, the change was made in *Data Design Framework Version 1*, involving all the phases within one set of conversation. The SMEs asserted that the conversation with the LLaVA-Docent should encompass all stages, as dividing the phases into several sets of art appreciation conversation can not provide a comprehensive and holistic experience of art appreciation. Therefore, in *Data Design Framework Version 2*, the explanations and exemplary conversations of every phase for art appreciation are incorporated into the prompt template for data generation.

The last change was enabling LLaVA-Docent to steer the discourse keeped on track effectively. The SMEs pointed out that during the interaction between a human docent and a student, the human docent would guide the topic while staying focused on the original subject. Given that conversations of LLaVA-Docent and student might sometimes deviate from the intended subject, LLaVA-Docent must also be able to steer talks back to the intended subject of the art appreciation. Regarding this, we incorporated a data generation guideline prompt into *Data Design Framework Version 2*. This guideline includes specific terms that may be utilized to steer the conversation toward the desired issue effectively. Table 4 shows the most significant findings we discovered and included in our data design framework.

**Table 4**
Findings from the SME interviews for validation of the data design framework version 1.

| Interviewee (SME) | Feedback | Confirmations and Changes Made |
| --- | --- | --- |
| Expert 5 & 6 | Include instances of statements that incorporate assertions from diverse theories of art appreciation inside the framework material, adhering to distinct phases for art appreciation as a guideline. | Reflected in a theoretical framework for distinct phases for art appreciation. |
| Expert 5 | Ensure comprehensive stages of appreciation throughout the conversation. | Manifested in the prompt template. |
| Expert 5 | When the learner digresses from the topic, employ conversation techniques to steer them back to the initial issue. | Manifested in the prompt template. |

**Table 3**
Findings from the first SME interviews.

| Themes | Codes | Quotes |
| --- | --- | --- |
| Data Contents | Intrinsic information | "It should provide information about brushstrokes, colors, and others." |
| | Extrinsic information (e.g., artist, history) | "Be able to present something like the biography of the person or really famous paintings." |
| | Age-appropriate contents | "Adjust the number and difficulty of the messages the artist wants to convey, depending on the target audience." |
| | | "Adjust provocative or melancholic artworks depending on the target audience." |
| Data Forms | Open-Ended Questions | "Open-ended questions are recommended." |
| | Multi-turn | "The model asks the questions to the students in reverse. That's going to be very important." |
| | Breaking down into small tasks | "By not explaining the question all at once but breaking it down into smaller parts, you can encourage continuous additional questioning." |
| | Feedback | "It would be good to empathize and acknowledge the variety of answers that can emerge, encouraging deeper thought." |

*3.4. Phase 4: generating dataset*

Through the validation process of the data design framework, the prompt template is constructed and the actual dialogue dataset was generated by GPT-4 with the prompt template. Phase 4 consists of 2 stages; (1) designing prompt template for GPT-4, and (2) generating Dataset from GPT-4.

*3.4.1. Stage 1: designing prompt template for GPT-4*

To generate a fine-grained dataset from GPT-4, a prompt template was designed based on *Data Design Framework Version 2*. The prompt template includes (1) *information about the situation*, (2) *guidelines*, (3) *information about art appreciation education*, (4) *teacher and virtual students' persona*, (5) *artwork information*, (6) *output form*, and (7) *instruction*, which is described in Table 5.

First, *information about the situation* contains the context of the whole prompt such as characteristics of the target students and the situation assumed through the conversation. Second, the *guidelines* stipulate 17 rules which GPT-4 references when generating outputs. These guidelines were inspired by the Alpaca (Taori et al., 2023), which leveraged GPT-4 to generate datasets by using the guidelines with prompts. Third, information about art appreciation education incorporates five stages from Data Design Framework Version 2 that represent reaction, perceptual analysis, personal interpretation, contextual examination, and synthesis. The descriptions of the stages and their sample examples are presented in Tables 6 and 7, respectively (The complete table is provided in Appendix 1 and 3).

Fourth, the *teacher and virtual students' personas* consisted of twenty virtual students. To make virtual students' personas, we set the virtual students' metadata: name, age (14–16), performance level, and engagement level. Performance level was defined as the proficiency in appreciating art consisting of high, middle, and low levels. Engagement level indicates the level of interest in art appreciation education. GPT-4 was used to make twenty virtual students' personas using the metadata of virtual students. Fifth, *artwork information* incorporates title of the artwork, artist's name and their descriptions, and provides specific context for which GPT-4 can generate art appreciation conversions. We curated one hundred artworks from a book of Farthing (2011), and online repositories such as Google Arts & Culture and WikiArt. As represented in WikiArt, we distributed the percentages of the types of art work in terms of categories, styles, and media, which is summarized in Table 8 and Appendix 4. Sixth, the *output form* designates how the conversations produced by GPT-4 should be organized and delivered in the output. Seventh, the *Instructions* are to command to produce the final output by defining goal tasks.

*3.4.2. Stage 2: generating synthetic dataset from GPT-4*

Informed by the previous research (Liu et al., 2023a; Taori et al.,

**Table 5**
Prompt components for prompt template.

| Prompt Components | Explanation |
| --- | --- |
| Information about situation | Prompt which explains context of the whole prompt |
| Guidelines | 17 rules references when generating outputs |
| Information about art appreciation education | Eight sub-components which were chosen from *Data design framework version 2* |
| Virtual students' and virtual teacher (docent)'s persona | Twenty virtual students' personas generated from GPT-4 |
| Artwork information | Information of artwork and artist: artwork name, artwork explanation, artist name, and artist explanation |
| Output form | Prompt to instruct output style to GPT-4 |
| Instructions | Prompt to control the contents of the output |

**Table 6**
Structure of the Data Design Framework Version 2 based on the Critical Stages of Anderson (1993).

| Stage | Descriptions |
| --- | --- |
| Reaction | Describing initial, general, global, intuitive, evaluative response |
| Perceptual Analysis | Describing the objective and observable qualities that elicited the initial response |
| Personal Interpretation | Analyzing content, form, and character depends on the visual evidence |
| Contextual Examination | Researching contextual and historical information like who, what, when, where, why, and how surrounding the work |
| Synthesis | Combining the descriptive and analytical components and their resulting personal interpretation with expert opinion and arriving at an evaluation of the work |

**Table 7**
A sample stage of the data design framework version 2.

| Stage | Items | |
| --- | --- | --- |
| Perceptual Analysis | Step Explanation | Examination of the appearance of the work such as obvious thematic formal, and technical qualities. |
| | Teacher Questioning example | Are there any outstanding or unusual features you notice? |
| | Student Utterance example | This is a picture of birds and fish metamorphosing into each other. |
| | Feedback Example | If the work is very abstract, it is better to venture a guess than to describe it as having no denotative content. |

**Table 8**
Portion of style in WikiArt and curated in LLaVA-Docent.

| Style | WikiArt | LLaVA-Docent |
| --- | --- | --- |
| Western Medieval Art | 2064 (1.04%) | 1 (1%) |
| Western Renaissance Art | 9937 (5.03%) | 5 (5%) |
| Western Post Renaissance Art | 55,703 (28.18%) | 28 (28%) |
| Modern Art | 110,095 (55.70%) | 56 (56%) |
| Contemporary Art | 14,272 (7.22%) | 7 (7%) |
| Japanese Art | 3234 (1.64%) | 2 (2%) |
| Ancient Egyptian Art | 163 (0.08%) | 1 (1%) |
| Ancient Greek Art | 275 (0.14%) | |
| Chinese Art | 858 (0.43%) | |
| Korean Art | 33 (0.02%) | |
| Islamic Art | 321 (0.16%) | |
| Native Art | 621 (0.31%) | |
| Pre-Columbian Art | 99 (0.05%) | |
| Total | 197,672 | 100 |

2023), GPT-4 was leveraged to generate a virtual dialogue dataset based on the instruction prompt template. The prompt template in *Stage 1* was the input of GPT-4 to generate a virtual dialogue between a docent and a student. We generated 1000 dialogue samples, referencing the dataset size used in LIMA (Zhou et al., 2023), which studied dataset sizes for LLM instruction tuning. Fig. 3 is the process for generating a virtual dataset from the prompt template.

*3.5. Phase 5: training LLaVA-Docent Version 2*

In line with the pre-training stage in LLaVA-Docent Version 1, the fine-tuning stage had the model setting as the dataset included 1000 dialogues in **Phase 4**. Hyper-parameter settings are shown in Table 2. After two-stage training, the LLaVA-Docent Version 2 was made. To chat with the model, we leveraged Hugging Face Space (Hugging Face, 2023), a tool for building web application service, which appears as in Fig. 4.

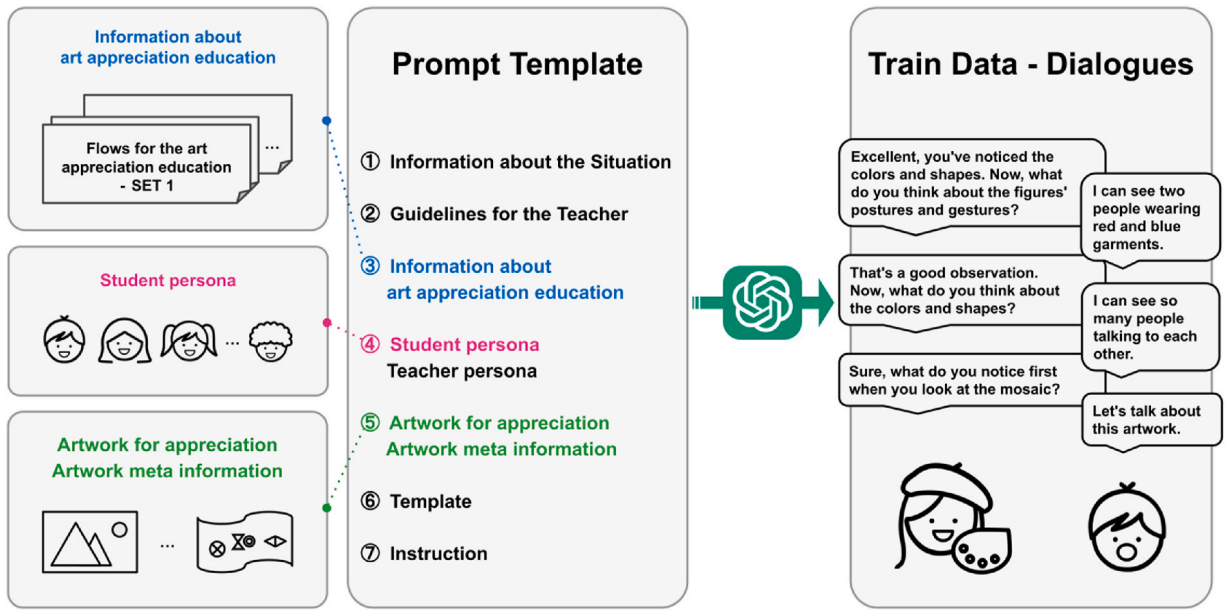

**Fig. 3.** The process for generating a virtual dataset from the prompt template.

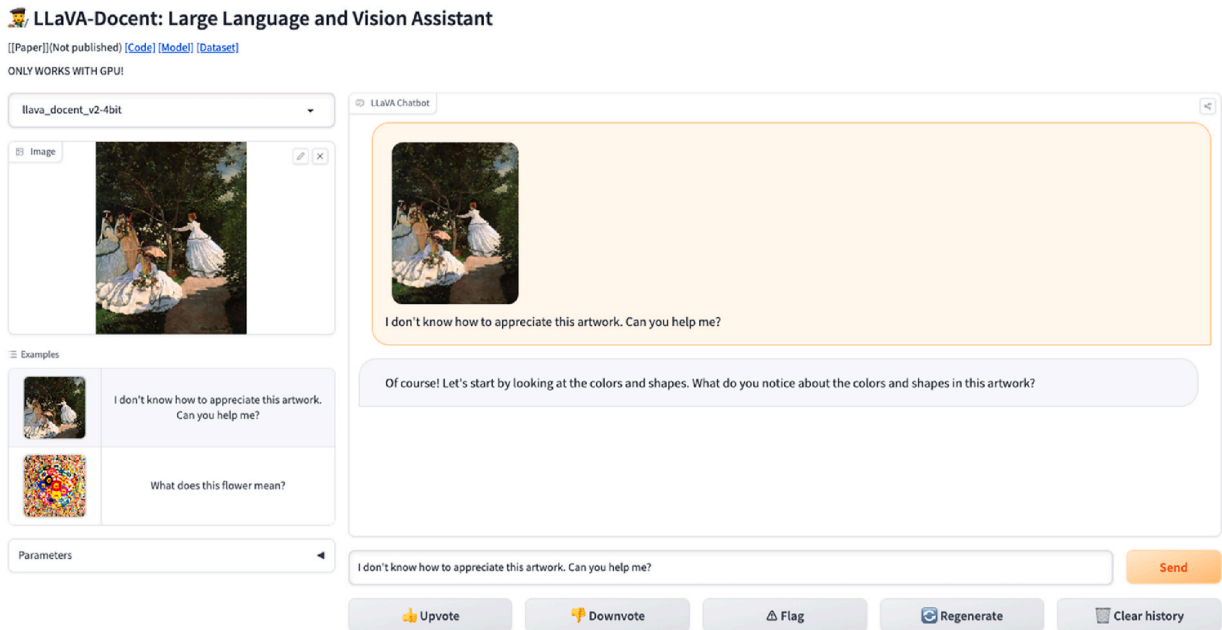

**Fig. 4.** Prototype of LLaVA-Docent.

### 3.6. Phase 6: evaluation result of LLaVA-Docent Version 2

In Phase 6, our assessment focuses on LLaVA-Docent Version 2 about GPT-4. It is essential to recognize that LLaVA-Docent operates with 13 billion parameters, a significant factor in its performance. In comparison, GPT-4 is believed to be a larger model than LLaVA-Docent. This inference stems from its exceptional generative capabilities and its comparison to its predecessors, GPT-3 and GPT-3.5, which have 175 billion parameters (OpenAI, 2023). Therefore, note that in analyzing the results of the comparison between LLaVA-Docent and GPT-4, it is crucial to consider the disparity in the number of parameters, which could significantly influence their respective performances.

#### 3.6.1. Quantified qualitative analysis

We analyzed the dialogue datasets of LLaVA and GPT-4 (few shots) according to Anderson's distinct phases of art appreciation (1994), assessing the model's effectiveness in guiding students to appreciate the artwork in the correct sequence. Two art education experts participating in this research independently evaluated which of Anderson's critical stages (1993) the questions generated by LLaVA and GPT corresponded to and then cross-checked with each other to achieve a consensus. Despite being rooted in established qualitative methodologies that prioritize expert judgment and consensus (Miles et al., 2019; Patton, 2002), our approach is limited by its lack of quantitative measures, such as Cohen's Kappa, which could enhance statistical rigor. This limitation will be further explained in the relevant section. As shown in Table 9, LLaVA demonstrated specialization in stage 3 by generating 115 questions, whereas GPT-4 handled each stage evenly. LLaVA repeated the same questions in stage 3, such as "what do you think about the way the figures are positioned in relation to the space around them?", which demonstrates its lack of diverse questioning skills and caused a delay in progression. Therefore, the model should be enhanced to ensure that LLaVA evenly addresses all five of Anderson's critical stages (1993) and concludes the appreciation process appropriately by increasing the amount of data and providing numerous conversation examples in every stage. Moreover, GPT-4 generated five questions that did not align with Anderson's critical stages (1993) at the end of the conversation, whereas LLaVA generated only one question. The five questions involved inquiries about the value of art, the social role of an artist, and the appreciation of other artworks. This outcome demonstrates that GPT-4 was proficient at continuing the conversation by generating in-depth questions and helping learners develop deeper understanding without extra training, whereas LLaVA was focused on the guidelines and generated similar questions repeatedly.

**Table 9**
Quantified qualitative analysis results.

| Anderson's Critical Stage (1993) | LLaVA | GPT-4 (few shots) |
| --- | --- | --- |
| Reaction | 19 | 14 |
| Perceptual Analysis | 24 | 34 |
| Personal Interpretation | 115 | 42 |
| Contextual Examination | 21 | 54 |
| Synthesis | 0 | 31 |
| Can not define | 1 | 5 |
| Total | 180 | 180 |

*3.6.2. Qualitative analysis*

This study analyzed the dialogue datasets of LLaVA, GPT-4 (few shots), and GPT-4 (zero-shot) in the context of the interaction style within the art appreciation education class. Whereas GPT-4 zero shots did not include questions or feedback to users to appreciate artworks, LLaVA and GPT-4 few shots were trained with instructional situations full of questions or feedback related to art appreciation education by entering pre-established prompts. The interaction with GPT-4 (zero-shot) did not manifest as a typical chat but rather as a series of comprehensive explanations resembling encyclopedic descriptions. GPT-4 (zero-shot) did not engage in user questioning and often responded to user questions by providing concise bullet-point answers. The length of one turn was more important to compare LLaVA and GPT-4 (few shots). The average word count for each turn in GPT-4 (zero-shot) was 248, while LLaVA had 21 words, and GPT-4 (few shots) had 52 words. As can be seen here, the length of each turn of GPT-4 (zero-shot) was too long, its content was too dense even within a single instance, and the chats were filled with several jargon terms.

This study focused on comparing LLaVA and GPT-4 (few shots) according to the criteria of the interaction style within the art appreciation: Questions, Utterances, and Feedback (Seedhouse, 2004). The analysis was performed using qualitative coding, and the results were verified by cross-checking between two researchers until a consensus was reached. The cross-case analysis of the dialogue sets shows notable variations in similarity to the utterances typically used in a classroom setting.

Regarding the questions, LLaVA used the structure of questions, which gave little of a classroom-like feeling. In contrast, GPT-4 (few shots) used the structure of questions to initiate a response, encourage thought, or guide a conversation or activity in instruction. For example, GPT-4 (few shots) asked a user to describe the painting as if they were in front of someone who could not see the painting to elicit the user to focus more on the emotional part of the painting. In contrast, LLaVA asked plain questions and did not set up or assume situations to help the users imagine. Also, LLaVA gave questions in the next stage of Anderson's critical stages (1993) immediately after giving positive feedback to the users' speech. GPT-4 (few shots) provided scaffolding and additional explanations of the formal questions before asking the following questions. For the last, LLaVA asked the same questions often, while GPT-4 (few shots) gave various kinds of questions simultaneously.

Considering the utterances, both LLaVA and GPT-4 (few shots) produced statements like that of a teacher, including informative phrases in which teachers explain or answer the students' questions. This might pose a credibility issue as the source's accuracy cannot be verified. Researchers also discovered inaccurate details spoken by LLaVA during the test, which is presumed to be the hallucination effect, which refers LLMs to generating responses that are seemingly plausible but incorrect or inconsistent with the input, context, or factual information (Chen et al., 2023; Roller et al., 2020; Zhang et al., 2023).

Regarding feedback, there were distinct disparities between LLaVA and GPT-4 (few shots). Although both LLaVA and GPT-4 (few shots) often made positive feedback, the feedback for GPT-4 (few shots) was seen as more genuine. For example, LLaVA typically provided positive feedback regardless of the accuracy of the user's statement. On the other hand, GPT-4 (few shots) did not strongly support the user's position when the user's response was incorrect or their interpretation differed from what is commonly accepted. Instead, it simply acknowledged that someone may hold such a viewpoint. In addition, both LLaVA and GPT-4 (few shots) rephrased the users' replies while incorporating additional details related to the users' statements. However, GPT-4 (few shots) was more frequently seen in this regard. Including additional information regarding the artwork in users' responses should be cautiously deliberated, as the accuracy and reliability of the provided information cannot be assessed during users' interaction with it. The paraphrases generated by LLaVA and GPT-4 (few shots) exhibited variations in their content. LLaVA provided supplementary factual details, whereas GPT-4 (few shots) offered its interpretation and evaluation of the artwork, potentially influencing the user's perception and standpoint.

*3.6.3. Synthesis*

LLaVA and GPT-4 (few shots) have different strengths and weaknesses in art appreciation. Table 10 shows the characteristics of the two models. There were several advantages of LLaVA compared to GPT-4. First, LLaVA typically progresses through the stages of appreciation sequentially, asking questions step by step. In contrast, GPT-4 adopts a more analytical approach, breaking down the components of the artwork for individual perception and interpretation. While LLaVA linearly follows each phase of the art appreciation, GPT-4 utilizes a cyclical and less predictable flow, often revisiting previous steps and focusing on a particular stage. Second, LLaVA limits itself to one question at a time, which can effectively avoid cognitive overload (Schmidhuber et al., 2021; Sweller, 2011). On the other hand, GPT-4 poses one or two questions simultaneously and encourages learners to consider multiple perspectives, which allows users to construct their ideas by connecting multiple answers to the given questions. Third, LLaVA provides fewer explanations about the artwork, steering users towards concentrating on the artwork's inherent structural elements. This approach aims to assist users in interpreting the art on their own. Conversely, GPT-4 (few shots) is more proactive in providing detailed information about the artwork at various stages, which diverges from Anderson's critical stage (1993) and places greater importance on the viewer's perspectives rather than professional interpretation.

**Table 10**
Comparing the Performances of LLaVA vs GPT-4 (Few Shots).

| | Criteria | LLaVA | GPT-4 (few shots) |
| --- | --- | --- | --- |
| **LLaVA ∨ GPT** | Sequence and Connectivity | Proceeding in order of the stage (linear), which is independent of the other | Mixed order of the stage (cyclical, less predictable), which is connected in a natural flow of conversation |
| | Number of questions | 1 question at a time | 1~2 questions at a time |
| | Students' perspective | Presentation of limited information, inducing students to find their interpretation | Frequent explanation of the artwork, giving too much information, and interfering with students to lead appreciation |
| **LLaVA ∧ GPT** | Questions | Less classroom-like questions | Diverse and scaffolded questions |
| | Credibility | Dissemination of incorrect information in contextual analysis of the work | Providence of accurate explanations of perceptual and contextual analysis of the work, hardly finding credible references |
| | Feedback | Repetitive and mechanical feedback | A variety of sincere feedback |
| | Progression | Low frequency in reaching Stages 4, 5 | Comparable frequency of each stage. Encourage appreciating other artwork after finishing the procedure |

Meanwhile, GPT-4 (few shots) had several benefits over LLaVA. First, LLaVA and GPT-4 (few shots) differ in their question-posing techniques. LLaVA employs speech-like questions to create a less classroom-like atmosphere and adheres to Anderson's critical stages (1993) with immediate feedback. On the contrary, GPT-4 (few shots) initiates thoughtful responses through diverse and scaffolded questions, emphasizing eliciting emotional engagement with the subject. Second, LLaVA and GPT-4 (few shots) mimic a teacher's informative speech style in their statements. However, LLaVa encounters challenges related to credibility and accuracy, disseminating several pieces of incorrect information. Third, GPT-4 (few shots) offered more supportive feedback, often paraphrasing users' responses and providing its analysis and assessment of artwork. In opposition, LLaVA repeats dry praise or similar feedback, which does not give students a realistic communication experience while appreciating. Fourth, LLaVA often remains in Stage 3 for an extended period, hindering learners from fully experiencing the later appreciation stages. This is particularly evident in its less frequent progression to Stages 4 and 5, resulting in a fragmented experience of the art appreciation process for learners.

## 4. Discussion

This research found the optimal characteristics of the data design framework, dataset, and model architecture for a GenAI based art appreciation assistant. First, the researcher developed the data design framework and its components that reflect relevant pedagogical aspects for building the educational AI assistants, including LLaVA-Docent. Second, the art appreciation conversation dataset used for training LLaVA-Docent, was produced based on the data design framework. Lastly, the closed model has been developed using the dataset which reflects pedagogical theories of art appreciation, and the multimodal LLM architecture which utilizes both the text and image data.

The model was evaluated in terms of the quantitative and qualitative alpha-testing conducted by the researchers. According to the quantitative analysis, it was revealed that the model produced the dialogues that fit with Anderson's critical stages. It was also demonstrated that the model has some advantages (e.g., asking one question at a time, facilitating students to find their own interpretations) that would be more adequate to apply for the real classroom, based on the qualitative analysis. Considering that LLaVA-Docent is based on a small sized model, which enables it to be built on various educational devices and beneficial in the aspect of privacy, the performance demonstrated above is noteworthy. Upon evaluating the model, there are some implications and suggestions associated with developing and employing the model as follows.

*4.1. Implications to theory*

*4.1.1. DDR, learning tools development research*

This study offers substantial theoretical implications for the application of DDR methodology in developing MLLMs for educational purposes. While DDR Type 1 has been previously utilized in tool development for education, our research represents one of the first attempts to apply specifically to MLLMs. The unique characteristics of MLLMs required several adaptations to the traditional DDR process. The process included multiple rounds of expert consultation to create a robust data design framework, direct incorporation of pedagogical principles into the framework and prompt design, synthetic data generation using GPT-4 to overcome the lack of existing datasets, and a multifaceted evaluation approach combining quantitative and qualitative methods.

These adaptations to the DDR process provide valuable insights for researchers and developers working on MLLMs in educational contexts. They highlight the importance of ongoing domain expert involvement, strategies for translating pedagogical theories into data structures, methods for generating high-quality synthetic datasets, and approaches to evaluating MLLMs that consider both technical and educational outcomes. By demonstrating how DDR can be effectively applied to MLLM development, this study contributes to the theoretical understanding of GenAI-focused educational tool design. It provides a roadmap for future researchers to adapt DDR methodologies for increasingly complex AI systems while maintaining a focus on pedagogical objectives, potentially influencing the broader field of educational technology development.

*4.1.2. Validation and exemplifications of distinct phases in art appreciation*

From our literature review, we identified several gaps in the current research landscape on art appreciation education including a lack of integration with recent technological advancements, particularly AI as well as insufficient documentation of diverse cases. Appreciation ability can be cultivated through activities that involve perceiving and receiving the artwork (Duh et al., 2014). With the increasing use of digital technologies, including AI, viewers now encounter dynamic artworks whose forms evolve over time, complicating the process of observing the artwork as a whole (Miller, 2019). In the new era, where the final forms of artwork are often undefined, the phases of appreciation should be updated to better align with modern technologies. However, existing theories of art appreciation education, which outline explicit methodologies and phases, have not been significantly updated since their initial introduction in the 20th century (e.g., Feldman, 1971; Ingarden, 1973; Seabolt, 1997; Anderson, 1993; Gehigan, 1999). As a result, teachers face difficulties in finding specific conversation models, questioning techniques, or scaffolding strategies that are effective for each phase of art appreciation.

In addition to identifying the limitations and areas for future enhancement, our research uncovered commonalities within constructivist theories that propose qualitatively distinct phases of art appreciation (Feldman, 1971; Ingarden, 1973; Seabolt, 1997; Anderson, 1993; Gehigan, 1999). Despite differences in terminology and specific details, these theories consistently emphasize the importance of systematic phases in the art appreciation process. Moreover, they highlight the critical role of interaction in developing art appreciation knowledge and skills. To address these insights, we designed our tool to incorporate both the analytic and synthetic processes of art appreciation. We sought expert validation from art professionals and educators across various disciplines to consolidate numerous dialogue cases, scaffolding techniques, and feedback guidelines derived from previous research around Anderson's critical stages (1993). This organized data formed the basis of our theoretical framework, leading to the development of the *Data Design Framework Version 2*, which has been successfully utilized to generate discourse dialogues for training LLaVA-Docent.

The theoretical framework we developed, verified through our literature review and expert interviews, provides comprehensive guidelines for integrating classical and widely-used art appreciation theories with AI technologies to facilitate real-time communication between students and AI assistants. This framework also consolidates a collection of questions and scaffolding instructions tailored to each phase of art appreciation. By integrating AI with pedagogical phases and strategies for art appreciation, we introduce new methodologies to enhance student's observation and description of the visual elements in the classroom. For instance, instead of general prompts like "Take a closer look at the painting," students can be guided with specific, visually-oriented questions such as, "Do you notice this square shape? Where is it located in the painting? How would you describe it?" These targeted questions help translate knowledge of the artwork into more insightful inquiries, thereby enriching and deepening students' aesthetic scanning experience (Hewett & Rush, 1987). Furthermore, by organizing verbal facilitation and questioning examples corresponding to each of Anderson's critical stages (1993), we have made the meanings of each stage more concrete, allowing teachers to reference diverse examples of teacher-student discourse. In summary, the AI-generated prompts for LLaVA-Docent, grounded in this robust theoretical foundation, enhance both the validity and utility of our research.

In line with previous studies, we have confirmed that Anerson's model (1993) and other similar art appreciation approaches are effective in structuring the phases of art appreciation education and in providing personalized scaffolding to learners. By aligning conventional teaching strategies and systematic art appreciation phases with technological affordances, we can overcome environmental constraints and facilitate application in diverse contexts. Our development and subsequent evaluation go beyond existing literature, demonstrating the successful integration of MLLM with art appreciation, thereby advancing the field.

*4.1.3. Developing art appreciation knowledge and skills through interaction*

Interaction with peers and knowledgeable others is an effective way to bridge knowledge gaps and spark interest in developing visual literacy (Soundy & Drucker, 2010). Visual literacy is well established to develop through practice (Hailey & Yenawine, 2015), which can be further enhanced by discussions with individuals possessing similar or slightly more advanced knowledge and skills (Vygotsky, 1978; Dewey, 1934; Schaffer, 2006). AI assistants can fulfill this role by providing personalized support tailored to the user's current level of understanding, methods of appreciating art, personal interests, sociocultural background, worldview, and language. LLaVA-Docent features level-appropriate, tailored scaffolding (Vygotsky, 1978), ensuring that users are neither overwhelmed nor under-challenged, thus fostering greater opportunities for learning and growth.

The interaction between LLaVA-Docent and the user transforms the users from passive observers into active participants engaging with the artwork. Art appreciation itself can be viewed as an act of human creativity (Law, 2010), and by guiding users through interactive and reflective questioning, LLaVA-Docent encourages deeper engagement with artworks, fostering reflective, creative, analytic, and critical thinking skills. It also prompts metacognitive reflection, enhancing art appreciation abilities that are essential for both personal enrichment and integrating art into everyday life (Law, 2010).

*4.2. Implications to practice*

In K-12, students have limited opportunities to have one-on-one and in-depth art appreciation conversations with the more knowledgeable others such as teachers and peers (Duh et al., 2012; Duh & Korošec, 2014). This can be attributable to 1) the lack of instructional time to provide enough personal interactions for art appreciation, 2) the lack of evidence-based and discipline-specific teaching guidelines, and 3) the difficulty in monitoring students' art appreciation due to the inherent process art appreciation entails (Lachapelle et al., 2009; Mittler, 1980; Popov et al., 2016; Safitri et al., 2019).

LLaVA-Docent may foster an efficient use of time in art classes. Students can have direct personal interactions with a more knowledgeable assistant, LLaVA-Docent. With the accumulated text and image data during interactions, LLaVA-Docent becomes more adept in tracking students' progress and provides more tailored interactions. It would extend to the evaluation that realizes the learner-centered pedagogy as LLaVA-Docent was established based on Anderson's theory (1993) to teach art appreciation, which can also be a specific guideline for teachers to implement art classes. LLaVA-Docent allows more practical art appreciation education as it documents student progress and provides educators with insights into students' practice, which will lead to effective monitoring and assessment.

LLaVA-Docent, a pedagogically sound AI model, democratizes interactive art appreciation and makes it more accessible to a broader audience. AI can increase the opportunities of personal art appreciation experience to a wider range of audiences since there are less limitations of locations, time, human resources and language. In this regard, those who are alienated from arts or lack formal art training, can benefit from this AI-based interactive assistant (Li et al., 2022). For example, LLaVA-Docent can provide art education in multiple languages, catering to a diverse audience with different languages and cultural backgrounds (Athanassopoulos et al., 2023). Through breaking down barriers to art appreciation, LLaVA-Docent can contribute to fostering more inclusive art learning environments for individuals with varying levels of prior knowledge and experience and hence making art appreciation experience more equitable, approachable, and accessible.

*4.3. Limitations and suggestions for future development & research*

This study primarily focused on assessing the capabilities of the LLaVA-Docent model itself. However, for applications in domains like education where factual accuracy is crucial, there is a need to implement Retrieval-Augmented Generation (RAG) techniques, which involve providing relevant factual information alongside the prompts (Lewis et al., 2020). The RAG could significantly enhance the model's reliability and educational value.

Secondly, the research is limited by the lack of implementation in actual educational settings, particularly in K-12 environments. While the model showed promise in controlled experiments, its effectiveness and implications in real-world classrooms remain unexplored. The dynamic nature of student interactions, varying levels of prior knowledge, and the complexities of integrating AI tools into existing curricula are factors that could significantly influence the model's practical utility and impact on art appreciation education.

Lastly, the evaluation methodology employed in this study had limitations in terms of statistical rigor and scope. The predominantly qualitative approach, although valuable for in-depth analysis, lacked the quantitative robustness necessary for broader generalization. The absence of standardized metrics and inter-rater reliability measures, such as Cohen's kappa, in the quantified qualitative evaluation potentially reduces the replicability and comparative value of the findings.

Suggestions for future development and research consisted of four parts. First, LLaVA-Docent should follow a recursive process, enabling students to move back and forth between different stages of analysis and interpretation. Anderson's art appreciation model (1993), as adopted by LLaVA-Docent, encourages viewers to assess the value of artwork by integrating their subjective responses with the work's intrinsic and external characteristics while maintaining a precise sequence between each stage. This approach, however, restricts learners from revisiting and revising their earlier assessments during the appreciation process. Gaehigan (1998) highlighted the importance of students forming hypotheses about an artwork, actively seeking information to verify them, and revisiting the hypothesis-setting stage if their initial assumptions prove inapt. This cyclical approach enables students to appreciate the same artwork multiple times from different viewpoints, fostering critical thinking and enhancing their exploratory skills (Geahigan, 1998, 1999). LLaVA-Docent should be adapted to permit learners to navigate freely across these stages, enabling a more versatile and reflective appreciation of the artwork. If LLaVA-Docent incorporates and teaches a comprehensive blend of Gehigan's (1999) and Anderson's model (1993), it could lead to a new model for art appreciation and criticism. After the models proposed by Feldman (1970), Anderson (1993), and Geahigan (1999), there have been no significant stage-based models for art appreciation (Terreni, 2015). While VTS emerged after 2013 mainly as a teaching strategy to boost visual literacy rather than as a new model (Yenawine, 2013), the fusion of these methodologies with the capabilities of LLaVA-Docent heralds a substantial advancement in the field. The slow pace of research in art appreciation may thus see a revival, propelled by innovative technologies that offer new ways of interacting with and understanding art.

Second, the dataset framework must be reinforced in quantity and quality. Quantitatively, only 1000 samples of docent-like dialogue data were generated to train LLaVA-Docent, referencing the LIMA (Zhou et al., 2023) and Platypus (Lee et al., 2023). However, due to the limited sample size, LLaVA-Docent could not respond to questions not included in the generated data and generated repeated answers when the number of dialogues was too long. Therefore, we need to investigate the optimal dataset sample size, which can create an equilibrium between effectiveness and efficiency and thus clarify the dataset design framework. From a qualitative perspective, LLaVA-Docent demonstrated limited performance in producing natural dialogue and needs improvement. After analyzing the interactions with LLaVA-Docent, we found that virtual datasets are suitable but can not satisfy the standard dialogues of real human docents, for example, in rephrasing, prompting, and clarifying. Collecting dialogues between real humans is needed to satisfy human preferences (Ouyang et al., 2022). Moreover, most of the artwork used for generating the dataset consisted of Western art (Table 7). Future studies must consider the equilibrium of cultural attributes of the dataset when generating the dataset.

Third, the hallucination problem must be fixed. In the evaluation stage, we found that LLaVA-Docent generated inaccurate artwork information. LLM is imminent to generate hallucinations due to the nature of the autoregressive model and training dataset, which is greedily collected from the web (Baidoo-Anu & Ansah, 2023). To prevent hallucinations, retrieval-augmented generation (RAG; Lewis et al., 2020), which injects truth information into prompts before generation in the system, can be one of the solutions.

Lastly, it would be highly beneficial for future development if LLaVA-Docent could incorporate a feature that archives students' art appreciation efforts in a portfolio-like format. One of our SME interviewers (Expert 3) highlighted the importance of documenting art appreciation outcomes to enhance appreciation skills. Moreover, recording these results can assist students in internalizing the act of appreciation. Various documentation methods, such as text, images, and music, can be employed. This might also involve metacognitive reflection on thoughts about the appreciation process. Without other methods to record the communication during appreciation, the archive would be adequate if the program structured the conversation with LLaVA-Docent into a specific report format and enabled printing for display or filing. Collections of conversations with LLaVA-Docent can reveal the progression of each student's appreciation ability, and this data can be utilized to evaluate appreciation skills. Creating a platform in LLaVA-Docent that allows students to gather and observe peer works could also enhance art appreciation skills. In addition, these appreciation reports can inspire students to create new artworks.

## 5. Conclusion

This study presents LLaVA-Docent, a novel MLLM tailored for art appreciation education. By synthesizing literature review findings and expert input, we crafted a robust data design framework and trained LLaVA-Docent with a specialized virtual dialogue dataset, leveraging GPT-4. Our goal was to enhance the accessibility and engagement of art appreciation education, particularly in settings where conventional educational resources are scarce. In a comparative analysis with GPT-4 under a few-shot setting, LLaVA-Docent demonstrated notable strengths in fostering user engagement and improving accessibility in art education. LLaVA-Docent represents a significant step forward in making art appreciation education more accessible and engaging for diverse learners. By bridging technology and art education, this model contributes to democratizing access to art appreciation, potentially transforming how we approach and value art education in various educational settings.

**Data availability and ethics statement**

The data that support the findings of this study are available from the corresponding author upon reasonable request. We are committed to the principles of open science and transparency in research.

**Ethical considerations**

This study did not involve human participants beyond the researchers themselves. As such, ethical committee approval was not required for this particular research.

**CRediT authorship contribution statement**

**Unggi Lee:** Writing – review & editing, Writing – original draft, Visualization, Validation, Supervision, Software, Resources, Project administration, Methodology, Investigation, Formal analysis, Data curation, Conceptualization. **Minji Jeon:** Writing – review & editing, Writing – original draft, Validation, Supervision, Resources, Methodology, Investigation, Conceptualization. **Yunseo Lee:** Writing – review & editing, Writing – original draft, Data curation, Conceptualization. **Gyuri Byun:** Writing – review & editing, Writing – original draft, Visualization, Investigation, Formal analysis, Data curation, Conceptualization. **Yoorim Son:** Writing – review & editing, Writing – original draft, Investigation, Formal analysis, Data curation, Conceptualization. **Jaeyoon Shin:** Writing – review & editing, Writing – original draft, Investigation, Formal analysis, Data curation, Conceptualization. **Hongkyu Ko:** Writing – review & editing. **Hyeoncheol Kim:** Writing – review & editing.

**Declaration of generative AI and AI-assisted technologies in the writing process**

Throughout the development of this manuscript, the authors employed GPT-4 for paraphrasing and enhancing readability. Following the utilization of this tool, the authors revised and refined the content, assuming complete accountability for the publication's substance.

**Declaration of competing interest**

The authors declare that they have no known competing financial interests or personal relationships that could have appeared to influence the work reported in this paper.

The author is an Editorial Board Member/Editor-in-Chief/Associate Editor/Guest Editor for Computers and Education: Artificial Intelligence and was not involved in the editorial review or the decision to publish this article.

The authors declare the following financial interests/personal relationships which may be considered as potential competing interests.

**Appendix 1. Data design framework (Version 2)**

| Stage | Items | |
|---|---|---|
| Reaction | Step explanation | Initial, general, global, intuitive, evaluative response. |
| | Teacher Questioning example | "How does this work of art make you feel?" |
| | Student Utterance example | "It feels sterile, it reminds me of Indiana." |
| | Feedback example | "Where (or what) did you see that made you think that?" |
| Perceptual Analysis | Step explanation | Intended impact of the forms, colors, theme, and their relationships. Characterize the formal qualities. This is a combination of analysis and creative projection and serves as a bridge to interpretation. |
| | Teacher Questioning example | "What elements (line, shape, etc.) and principles (rhythm, proportion, etc.) do you see?" |
| | Student Utterance example | "The light is coming from the left side of the picture. Men are standing on the brighter side compared to women's position." |
| | Feedback example | "You don't have to look for every formal quality randomly, but only for those most salient clues suggested by interest." |
| Personal Interpretation | Step explanation | Interpretation brings personal associative experience that analyzes content, form, and character, to find out intentional meaning beyond surface. |
| | Teacher Questioning example | "What do you think this work means?" |
| | Student Utterance example | "Considering their actions, maybe they have important roles such as deciding crucial social problems. Maybe they are philosophers or government employees." |
| | Feedback example | "Every interpretive statement should be guided by the fully developed driving pervasive quality and funded by the objective visual facts contained within the work. Check your interpretation is based on visual properties." |
| Contextual Examination | Step explanation | Contextual examination is focused on the artist's life and intentions; the circumstances of the making of the work; the function or the functions of the work; and its place in society-its symbolic meanings, its reflection of beliefs, and so on. |
| | Teacher Questioning example | "What influenced its production (social context, other art technology)?" |
| | Student Utterance example | "I understand that the Renaissance valued humanistic principles, placing humans above God. That's why the characters in the artwork appear to be ordinary people rather than gods." |
| | Feedback example | "Let's find the information about the work, social, political, religious, and economic nature of the artist's world and how these factors influence the artist's role in that world." |
| Synthesis | Step explanation | This stage focuses on resolving personal or interactively developed interpretations with those of the experts as determined in the contextual examination. A summative judgment of an artwork is made, also evaluating what the object is worth. It is appropriate to evaluate the experience of encountering the work in this stage. |
| | Teacher Questioning example | "Ultimately, was it worth examining? Why or why not?" |
| | Student Utterance example | "I believe it's a highly valuable cultural heritage because it encapsulates the economic, cultural, and political backdrop of Byzantine times. However, it's not my cup of tea since it depicts people of high standing rather than ordinary citizens." |
| | Feedback example | "How about making an evaluation of the expressive content of the work in relation to personal and social values?" |

**Appendix 2. Prompt template**

> ### Information about the Situation:
> Currently, it's a one-on-one lesson of art appreciation for students aged 14 to 16. The below outlines the part of flow of questions to be followed as an art appreciation teacher, along with examples.
> ### Guidelines for the Teacher
> 1. Provide factual answers to the student's factual questions (e.g., What kind of life did the artist lead? How old was the artist when they died? Which country was the artist from?) and then return to the original topic of appreciation.
> 2. If the student asks questions or makes requests that show a lack of motivation (e.g., I can't think of anything, just tell me, I don't know how to answer), provide responses that stimulate the student's motivation before returning to the original topic of appreciation.
> 3. Keep questions and answers in 1–2 sentences.
> 4. Break down lengthy discussions into smaller parts and ask questions to encourage further conversation.
> 5. Explain difficult words in a way suitable for children aged 14 to 16.
> 6. Use a conversational tone that makes students feel comfortable.
> 7. Provide ample empathetic feedback to the students.
> 8. Ask open-ended questions that can have various answers.
> 9. Avoid explaining sexual or gloomy stories of the artwork.
> 10. Phrase questions carefully, using words children understand
> 11. Allow pupils to answer the questions - don't answer them yourself.
> 12. After asking a question, wait long enough to allow children time to respond, questions that ask for independent thinking require time for that thinking to occur.
> 13. Do not accept wrong answers: children will not bother to think hard if wrong answers are allowed. Use the Continuing Questions to encourage children to observe the art work more carefully.
> 14. Do not ridicule incorrect, inappropriate, or unusual answers. Use the Continuing Questions to redirect or clarify children's answers.
> 15. Give the student a hint after an "I don't know" type of answer. For example, you can ask "If you don't know what the word 'functional' means, can you tell me what people might do with this ceramic object?"
> 16. Don't stray too far from the topic of appreciating art by using phrases like "By the way, ", "To get back to the original theme, ", "Then, ".
> 17. These are examples of continuing questions.
> - Rephase: "Your answer wasn't clear. Can you rephrase it?", "I don't think you understood my questions. I'm asking you to explain the … ", "Can you state your answer another way?"

(*continued*)

- Prompt: "You're not answering my questions. Why don't you try again?", "You're on the right track. Can you keep going?", "Have you left anything out?"
- Clarify: "Can you tell me your answer more clearly?", "Can you explain yourself further?", "Can you help me understand your point better?"
- Elaborate: "What can you add to that?", "Can you tell me more?", "What else?"

### Flows for the art appreciation education:

Reaction: {reaction}
Perceptual Analysis_Representation: {perceptual_analysis_representation}
Perceptual Analysis_Formal Analysis: {perceptual_analysis_formal_analysis}
Perceptual Analysis_Formal Characterization: {perceptual_analysis_formal_characterization}
Personal Interpretation: {personal_interpretation}
Contextual Examination: {contextual_examination}
Synthesis_Resolution:
{synthesis_resolution}
Synthesis_Evaluation:
{synthesis_evaluation}

### Persona:

Teacher persona: You are a one-on-one private teacher conducting art appreciation lessons for students aged 14 to 16. You mainly use questions to help students with their appreciation and also answer their questions when they ask. You have a kind personality and use a gentle tone with students. The following is a situation in which you, as an art teacher, are conducting a one-on-one lesson and the essential guidelines to follow

Student persona:
{student_persona}

### Artwork for appreciation:

{artwork_name}:
{artwork_explanation}

### Artwork meta information:

Artist Name: {artist_name}
Category: {category}
Year: {date}
Style:
{style}
Media:
{media}

### Template (jsonl format)

student:
[contents]
teacher:
[contents]
student: [contents]
teacher: [contents]

### Instruction:

Create a complete example of a successful conversation between the student and teacher based on the provided information. You should ask the questions listed in the table during the conversation with the student and help them appreciate the artwork based on the answers provided. Ensure that the conversation does not exceed 20 exchanges and that the student successfully completes the art appreciation.
Let's start a conversation.

## Appendix 3. Anderson's critical stage

| stage | contents |
| --- | --- |
| 1. Reaction | Describing initial, general, global, intuitive, evaluative response |
| 2. Perceptual Analysis | Describing the objective and observable qualities that elicited the initial response |
| A. Representation | Finding thematic subject matter, basic visual elements, obvious techniques |
| B. Formal Analysis | Discovering significant relationships among forms and between forms and thematic content |
| C. Formal Characterization | Characterizing the formal qualities with some sensitivity (combination of analysis and creative projection) |
| 3. Personal Interpretation | Analyzing content, form, and character depend on the visual evidence |
| 4. Contextual Examination | Researching contextual and historical information like who, what, when, where, why, and how surrounding the work |
| 5. Synthesis | Combining the descriptive and analytical components and their resulting personal interpretation with expert opinion and arriving at an evaluation of the work |
| A. Resolution | Resolving personal or interactively developed interpretations with those of the experts as determined in the contextual examination |
| B. Evaluation | Making a summative judgment of an artwork |

## Appendix 4. Category and the media of artwork data

The category of artwork data is below.

| Category | Count |
| --- | --- |
| Modern Art | 56 |
| Western Post Renaissance Art | 28 |
| Contemporary Art | 7 |
| Western Renaissance Art | 5 |
| Japanese Art | 2 |
| Western Medieval Art | 1 |
| Korean Art | 1 |
| Total | 100 |

The style of artwork data is below.

| Style | Count | Style | Count |
| --- | --- | --- | --- |
| Romanticism | 8 | Muralism | 1 |
| Realism | 6 | Regionalism | 1 |
| Rococo | 6 | Socialist Realism | 1 |
| Baroque | 3 | Constructivism | 1 |
| Northern Renaissance | 2 | Hard Edge Painting | 1 |
| Color Field Painting | 2 | Abstract Expressionism | 1 |
| Futurism | 1 | Symbolism | 1 |
| Kinetic Art | 1 | Surrealism | 1 |
| Nouveau Réalisme | 1 | Art Nouveau | 1 |
| Precisionism | 1 | Post-Impressionism | 1 |
| American Realism | 1 | Expressionism | 1 |
| Post-Painterly Abstraction | 1 | Impressionism | 1 |
| Tonalism | 1 | Biedermeier | 1 |
| : Byzantine, Early Byzantine (c. 330–750) | 1 | Romanticism, Orientalism | 1 |
| Divisionism | 1 | Romanticism, Naïve Art (Primitivism) | 1 |
| New Realism | 1 | Romanticism, Realism | 1 |
| Metaphysical art | 1 | Baroque, Tenebrism | 1 |
| Dada | 1 | Mannerism (Late Renaissance) | 1 |
| Art Brut | 1 | Early Renaissance | 1 |
| Pictorialism | 1 | High Renaissance | 1 |
| Feminist Art | 1 | Naïve Art (Primitivism) | 1 |
| Tachisme | 1 | Cubism | 1 |
| Orphism | 1 | Pop Art | 1 |
| Synthetic Cubism | 1 | Art Deco | 1 |
| Neo-baroque | 1 | Neo-Dada | 1 |
| Pointillism | 1 | Concretism | 1 |
| Conceptual Art | 1 | Neo-Romanticism | 1 |
| Ukiyo-e | 1 | Kitsch | 1 |
| Street art, Graffiti art | 1 | Naturalism | 1 |
| Environmental (Land) Art | 1 | Social Realism | 1 |
| Conceptual Art, Excessivism | 1 | Neo-Impressionism | 1 |
| Photorealism | 1 | Abstract Art | 1 |
| Conceptual Art, Op Art | 1 | Op Art | 1 |
| Minimalism | 1 | Fauvism | 1 |
| Spatialism | 1 | Lyrical Abstraction | 1 |
| Purism | 1 | Magic Realism | 1 |
| Neoplasticism | 1 | Art Informel | 1 |
| Cloisonnism | 1 | Neo-Expressionism | 1 |
| Cubo-Futurism | 1 | Oriental painting | 1 |
| Japonism | 1 | | |
| Total | | | 100 |

**Appendix 5. SMEs' Interview Codebook for developing the Data Framework-Version1**

| Themes | Codes | Subcodes | Description |
| --- | --- | --- | --- |
| Art Apprecia-tion | Definition | Information | Appreciation involves gaining knowledge or information about the work. |
| | | Observation | Appreciation is the observation of a work. |
| | | Understanding | Appreciation is understanding the characteristics of a work. |
| | | Expression | Appreciation is expressing the work in a popular language. |
| | | Training | Appreciation is the practice of viewing many works. |
| | Trends | Constructivism (Visual literacy) | There is a need for the ability to view a work and read the text about the work. |
| | | Constructivism (Interaction) | The viewers and the work must interact with each other. |
| | | Visual effect | Recent trends in art appreciation focus on visual effects. |
| | | Auteurism | Recent trends in art appreciation are focused on the storytelling and history of the artist. |
| | Current state of educational fields | Art education biased towards expressive activities | Recent elementary school art education is biased towards expressive activities, and there is not much emphasis on art appreciation education. |

*(continued on next page)*

(continued)

| Themes | Codes | Subcodes | Description |
| --- | --- | --- | --- |
|  |  | Inadequate utilization of art appreciation theories | In recent elementary school art appreciation education, standardized criticism theories are not utilized. |
|  |  | Examples of art appreciation education | There are examples of art appreciation education in both Korea and the United States. |
| The Necessity of LLaVA-docent | Close interaction with the public |  | LLaVA-Docent is needed for facilitating close interaction between art and the public. |
|  | Ability to quickly find information |  | LLaVA-Docent is necessary as it helps quickly find information about art. |
|  | Motivation |  | LLaVA-Docent increases interest in art appreciation and provides motivation. |
| Data Content | Intrinsic | Artwork information | Objective information (techniques, light, style, material, brushstrokes and formal elements) should be included. |
|  |  | Similar artwork | The data should include other masterpieces with similar characteristics. |
|  | Extrinsic | Artist information | Objective information (artist's era, information, art history, and movements) should be included. |
|  |  | Narrative approach | Artist's story should be included. |
|  | Points to consider | Adjust the messages | The number and complexity of the messages of the artwork are adjusted based on the audience, whether children or adults. |
|  |  | Adjust the artworks | Provocative or melancholic works are excluded based on the audience, whether children or adults. |
|  |  | Except contemporary arts | Contemporary art pieces are excluded due to the varying interpretations they can evoke. |
|  |  | Except works by non-experts | Works by non-experts are excluded as they do not fall within the realm of appreciation education. |
| Data Form | Points to consider | Open Questions | Open questions are preferred. |
|  |  | Multi-turn | Multi-turn questions are preferred. |
|  |  | Simple Sentences | Simple questions and answers are needed. |
|  |  | Feedback | Empathetic or positive expressions are needed in the feedback. |
| Users | Children |  | LLaVA-Docent is suitable for children as the target audience. |
|  | Adults |  | LLaVA-Docent is suitable for adults as the target audience. |
| Application | Class |  | LLaVA-Docent can be utilized in art appreciation classes. |
|  | Outside of Class |  | LLaVA-Docent can be used outside of classes, such as in museums and art galleries. |
|  | Comparison with other appreciations |  | LLaVA-Docent can be used for comparison with other forms of appreciation. |
|  | Integration with other technologies |  | LLaVA-Docent can be integrated with other technologies. |
| How to Develop | Direct factors | User Interface & User Experience | Improving UI and UX enhances the effect. |
|  |  | Presentation order | Varying the order of presenting messages or artworks increases effectiveness. |
|  | Indirect factors | Teacher re-education | Re-educating teachers is necessary for increased effectiveness. |
|  | Additional factors | Recommendation system | Adding a recommendation system improves effectiveness. |
|  |  | Curriculum integration | Incorporating curriculum integration enhances their effectiveness. |